# TAACKIT: Track Annotation and Analytics with Continuous Knowledge Integration Tool

Lily Lee*, Julian Fontes[†], Andrew Weinert[‡], Laura Schomacker[‡‡], Daniel Stabile[§] and Jonathan Hou[||]
Lincoln Laboratory
Massachusetts Institute of Technology
Lexington, MA, USA
Email: [*leel, †julian.fontes, ‡andrew.weinert, ‡‡laura.schomacker, §daniel.stabile, ||jonathan.hou]@ll.mit.edu

*Abstract*— Machine learning (ML) is a powerful tool for efficiently analyzing data, detecting patterns, and forecasting trends across various domains such as text, audio, and images. The availability of annotation tools to generate reliably annotated data is crucial for advances in ML applications. In the domain of geospatial tracks, the lack of such tools to annotate and validate data impedes rapid and accessible ML application development. This paper presents Track Annotation and Analytics with Continuous Knowledge Integration Tool (TAACKIT) to serve the critically important functions of annotating geospatial track data and validating ML models. We demonstrate an ML application use case in the air traffic domain to illustrate its data annotation and model evaluation power and quantify the annotation effort reduction.

## I. Introduction

Geospatial track data represents the geospatial positions of objects in time. It is an important source of information on the behaviors of moving objects and has applications in defense, transportation, logistics, and environmental monitoring. The large size of geospatial datasets makes them amenable to machine learning techniques to analyze and predict object behaviors in a geospatial context.

Machine learning (ML) application development relies on these key elements: ML algorithms, computational resources, and data used to train the ML models. According to some estimates, collecting, curating, labeling data, and iterations with training and validation of the ML model optimization, takes on average 80% of the effort of ML application development [1]. These datasets must be labeled with classes/tasks relevant to development goals. Example tasks for ML development could be detecting cats from images, bikes from videos, or classifying geospatial tracks as pedestrians or cars. In data domains such as images and videos, there are existing tools to provide annotation functionalities, such as Visual Object Tagging Tool [2] and Computer Vision Annotation Tool [3]. ArcGIS and MapBox are two the most commonly used software tools in the geospatial data domain. Both tools have the capabilities to ingest geospatial images and geographic features, include track data, in some standardized formats. Users are able to visualize, interpret, and annotate geospatial data. However, the primary output of these tools are analysis and reports, not annotated truth data for use in ML development. As of the writing of this paper, there are no readily available ML-centric geospatial track data annotation tool in the open-source community or accessible commercial services.

One way to tackle the significant effort required to provide an ML-ready dataset is to integrate machine learning and/or analytics capability in the annotation and data curation process to prefilter and pre-label data. The automated labels can be verified by humans to ensure the quality of input data for ML algorithm/model training. Snorkel [4] and Augmented Annotation [5] are examples of such efforts. Snorkel is a programming language that provides users with the capability to apply rules-based techniques to quickly sort through and label datasets for ML development. The Augmented Annotation tool applies visual tracking and object detection techniques to annotate video data for objects of interest and assign class labels. Human annotators can verify the automated labels for quality assurance before iterating through ML training and validation steps. In both cases, the integration of automation in pre-labeling and verification by human experts can significantly reduce the level of effort required in developing ML applications.

This paper introduces Track Annotation and Analytics with Continuous Knowledge Integration Tool (TAACKIT), which is the first tool to provide the capability to annotate geospatial track data and enable the integration of annotation and model validation to improve efficiency in geospatial ML application development process. Our goal in developing TAACKIT is to not only provide the annotation capability but also to minimize effort on ML-ready data annotation and management process by integrating ML/analytics into the data management process. We describe TAACKIT's implementation and demonstrate its capabilities through a use case in analysis and analytics development in air traffic surrounding an airport.

## II. Geospatial Annotation & analytics criteria

We begin with considering the criteria that a geospatial annotation and analytics validation tool needs to fulfill.

### 1) Adaptability to geospatial track data formats

Geospatial track data can be highly varied in formats attributable to the variety of sensors used for data collection, and the many methodologies employed for standardization. As a result, TAACKIT needs to enable user definitions of input data.

### 2) Adaptability of annotation goals

ML application development goals and annotation label sets are dependent on user objectives. TAACKIT needs to enable users to define their own label sets.

### 3) Enabling ML model iteration & validation process

ML development processes often require multiple iterations of data annotation, model training, and validation. Leveraging an ML model in the annotation process can

DISTRIBUTION STATEMENT A. Approved for public release. Distribution is unlimited.
This material is based upon work supported by the Department of the Air Force under Air Force Contract No. FA8702-15-D-0001. Any opinions, findings, conclusions or recommendations expressed in this material are those of the author(s) and do not necessarily reflect the views of the Department of the Air Force.

significantly improve the efficiency of the annotation and model validation process.

TAACKIT has been implemented with the goals of adaptability in data formats and annotation tasks, efficient user interface, and enabling ML model performance validation.

## III. TAACKIT IMPLEMENTATIONS

TAACKIT is built on Cesium, an open-source 3D geospatial data visualization platform [6]. The TACCKIT software architecture illustrated by Fig.1, comprises of a React frontend that facilitates user interactions with geospatial data, enabling functionalities such as uploading, visualization, filtering, and assigning annotations. Integration with Cesium provides a comprehensive visualization of the data on a globe while providing a map server. To accomplish this, the front end sends requests to a FastAPI gateway, which invokes designated backend microservices that interact with a PostgreSQL database for user account management, and the storage, updating, and deletion of data or user annotations. TAACKIT is deployed as a web service with user authentication allowing it to serve multiple geospatial track annotation projects.

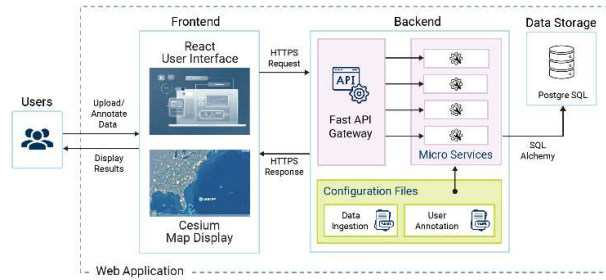

Fig. 1. TAACKIT Software Architecture

The TAACKIT track database contains the following tables:

1. Track position table, which contains the temporally ordered 3D points of geospatial tracks
2. Track annotation table, which contains properties about the entire track such as number of points, start and end times, number of annotations, etc.
3. Annotation table, which stores the annotation sets associated with each project
4. Annotator table, which stores both human annotator user IDs and roles and ML model names and iteration IDs.

TAACKIT's track position table contains a standard set of columns that are common to 3D geospatial tracks such as timestamp, latitude, longitude, altitude, and track ID. Additional columns can be defined by user as needed. Geospatial track data comes in many different formats, and are often custom-defined for the sensors producing the track data. As a result, naming conventions for standard track position fields are often different. To enable users to upload custom track position data formats into TAACKIT, a track position data format YAML file is defined by the users and read by TAACKIT to parse the position data.

Geospatial track annotation tasks are tailored to the ML application development goals. An ML application classifying ground-based traffic may need to label tracks as car, bicycle, or pedestrian, whereas an application on air traffic analytics may want to use helicopter and plane labels. Users define the annotations for each project through a YAML file, to be read by TAACKIT and used for display in the annotation interface.

TAACKIT provides the capability to export annotations for use by ML model training and or testing processes. Each annotation is associated with an annotator, thus enabling downstream evaluation of annotators.

While it is possible to manually annotate all the data necessary for ML development before commencing model selection, training, and testing, it can be more efficient to train models with a labeled sub-dataset, run inference on the rest of the dataset, and then validate these preliminary labels. A model trained on a subset of data may not be highly performant but could expedite annotating remaining data by eliminating the time-consuming task of individually annotating each track manually. TAACKIT accommodates this annotate-infer-validate workflow by allowing users to ingest external annotations which can then be reviewed and updated in bulk processing actions.

The ingestion of annotations is enabled through a user-defined YAML file describing the annotation algorithm and annotation labels. Once ingested into TAACKIT, users can leverage its query capability to locate tracks with labels of a particular characteristic, such as having class X produced by algorithm Y. Users can then batch-verify the queried tracks by reinforcing or correcting annotation labels.

## IV. TAACKIT USE CASE SCENARIOS

In developing an ML application for geospatial track data, an ML expert would need the following capabilities: sensemaking—the ability to verify the relevance of the data to development goals, annotating track data, ingesting external annotations, and visualizing tracks with corresponding annotations to verify for correctness, and iterating the ML training and testing process.

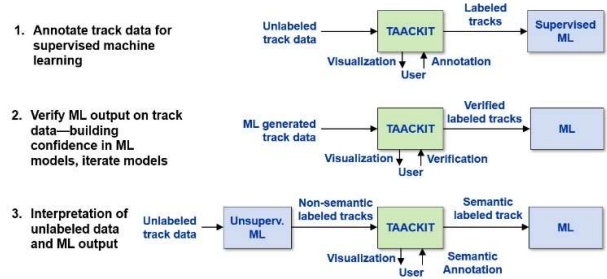

Fig. 2. TAACKIT use case scenarios: 1) Manual annotation of tracks, 2) Validation and iteration of ML model in development, 3) Data exploration and discovery in ML development

We demonstrate through use case 3 of Fig. 2 how TAACKIT functions in a geospatial track analytics development process by enabling exploration of unlabeled data to frame the analytics tasks, annotating tracks, and iterating analytics development while optimizing ML processes by increasing confidence and reducing the level of effort required in analytics development.

## V. USE CASE DEMONSTRATION

Among the many sources of geospatial track data, air traffic behavior is a significant domain with many analytics needs. We demonstrate the utility of TAACKIT in a use case involving understanding the air traffic behavior near an airport. An example of applications requiring an in-depth understanding of airport air traffic the NASA's Advanced Air Mobility (AAM) National Campaign, which aims to integrate new AAM vehicles into the airspace. The Air Traffic Management eXploration (ATM-X) project aims to enable autonomous aircrafts to operate safely within airport traffic patterns, under which the Traffic Pattern Intent Prediction (TPIP) project focuses on characterizing and predicting traffic patterns at small airports. OpenSky Network [7] is a global open-source air traffic surveillance network that provides real-time and historical aircraft position data. OpenSky Network-based datasets have supported aviation safety technology and standards development, such as [8]. To support the TPIP study on traffic patterns at small airports, a dataset that includes 11 million position updates in the entire month of May 2024 at multiple small airports was collected from the OpenSky Network. The criteria of 120 nautical miles of the airports and below 1500 feet above airport elevation were used to include air traffic behaviors specifically relevant to airport runways and exclude transiting airplanes.

Airplane behaviors near an airport can be generally categorized as transiting through the air space (usually at higher altitudes), taking off from the airport, landing at the airport, or a behavior known colloquially as touch-and-go. Touch-and-go is generally performed by novice pilots practicing takeoff and landing without stopping at the airport. Airplanes that transit above airports do not interact with the airport runways or exhibit any of the behaviors of interest, thus are excluded from the data set. The characterization of these behaviors is dependent on the geospatial locations of the planes and their temporal behaviors over time. The development of track analytics to classify air traffic tracks by behavior exemplifies the common challenges related to developing ML models for geospatial track data domain: large data volume, lack of truth data, and the need for validation of analytics in the optimization process. One possible solution is to upload all the air traffic tracks into our track annotation tool, TAACKIT, and annotate each track for behavior attributes. However, this is a highly labor-intensive effort and does not scale well for large datasets. We will use a small-scale demonstration of leveraging TAACKIT and the integration of unsupervised and supervised machine learning techniques to speed up the acquisition of truth data and the development of high-performance behavior classification model.

Unsupervised ML algorithms are practical solutions to address the need for analytics in large volume of unlabeled data. Supervised ML algorithms are often higher in performance than unsupervised techniques, but the need for labeled truth data is often daunting in large data volumes. Hence, we use unsupervised ML to bootstrap the initial truth data, then apply supervised ML to gain performance. Kmeans [9], an unsupervised ML algorithm, is used as a bootstrapping step to provide the initial rudimentary classification capability. For the purpose of this demonstration, we have chosen support vector machine (SVM) [10] as the supervised ML algorithm whose results that we will share in the paper, though many other multi-class classification models, such as multi-layer perceptron and random forest, could be used.

For the purpose of this demonstration, we choose airport code KARB as the airport of interest for analysis due to its small size and thus manageable number of tracks for demonstration[2]. A total of 1781 unique tracks and over 2 million position observations are included in this dataset.

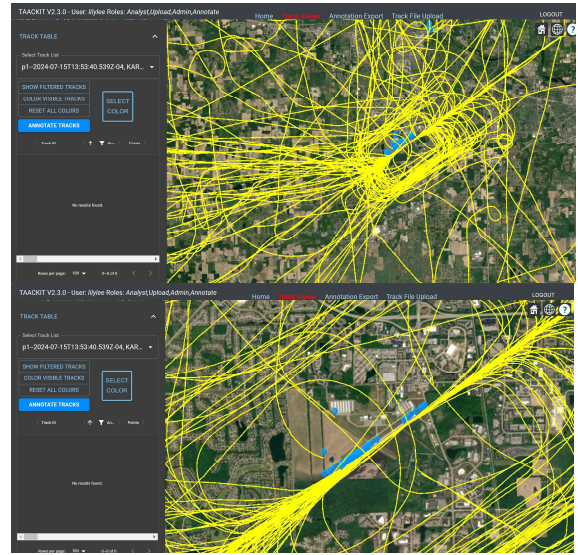

Fig. 3. Top: a wide area view of airplane tracks in the vicinity of KARB airport where the oval loops of the main runway are clearly visible. Bottom: a zoomed-in view of traffic directly over the airport. The blue pins are the beginning of tracks. The majority of track beginnings are on the main runway, with sporadic tracks on the un-named grassy runway.

### A. Data Exploration and Preprocessing

To ingest the OpenSky Network dataset into TAACKIT, we specify the geospatial track data format in a YAML file allowing TAACKIT to parse the data and visualize the tracks. Once the tracks are uploaded into TAACKIT, a user may explore the data with enhanced comprehension provided by overhead imagery of the area of interest. Fig. 3 shows a wide-area view of the airport with air traffic behaviors preceding landing/takeoff, and a zoom-in view of the behaviors directly related to landing and takeoff. Using the earth surface image context, a user could conclude that air traffic behaviors are dependent on the runway on which the traffic follows.

A number of preprocessing steps are needed to reduce the quantity of data for annotation and behavior analytics development. Airplane tracks are segmented to short distance ranges (8 km) surrounding the airport runways such that each track segment would exhibit a coherent behavior.

For example, if an aircraft that took off from the airport practiced a number of touch-and-go maneuvers and then landed, the aircraft track would be divided into multiple segments each with a single behavior class. Airport runway locations are detected using principal component analysis (PCA) on track positions. Track segments are classified by the runway using the angular distance between the runways and the average direction of the track segment. The runway classifications results are verified through TAACKIT. While runway classification is not strictly necessary to classify landing, touch-and-go, and takeoff behaviors, the behavior of the aircraft following these do depend on the runways. Additionally, visualizing tracks by runway improves interpretability for the human annotators.

*B. Air Traffic Behavior Classification*

Track segments annotated with behavior attributes are needed to train supervised ML algorithms such as SVM. Instead of manually labeling each track segment, we are using Kmeans, an unsupervised ML technique, to bootstrap the annotations. The landing, touch-and-go, and takeoff behaviors are largely characterized by the vertical velocity of the aircraft. We have chosen to capture the information using a 5-bin histogram of the vertical velocities of the individual track segments, where the velocities are estimated by differencing sequential positions. The histogram features are clustered using Kmeans where k=3 and initialized nominally with values corresponding to landing [0.5,0.5, 0,0,0], touch-and-go [0, 0.3, 0.4, 0.3, 0], and takeoff [0,0,0, 0.5, 0.5]. The Kmeans cluster assignments for track segments are ingested into TAACKIT for verification by human annotators.

To facilitate the iterative development of the track behavior classifiers and validation of the models, we split the track data into 4 sets with approximately 270 tracks per set. Sets 1, 2, and 3 used in training successive behavior classification models, and set 4 as the validation dataset. Human annotators were tasked to verify behavior class assignments on set 1 resulting from Kmeans clustering. Annotators used the query capability to visualize tracks with one behavior class assignment on one runway, as shown in Fig 4, then batch-annotate all tracks with the correct behavior. The effort required to batch-annotate a large number of tracks is identical to annotating a single track. Only the misclassified tracks (57 out of 271) needed individual annotations. Truth labels for data set 4 are similarly verified and used to evaluate all models. Table 1 compares all classifier performances.

The verified truth labels from set 1 are then used to train a multi-class SVM and evaluated using data set 4. The performance of SVM v1 is significantly improved compared to Kmeans, see Table 1 rows 1 and 2. We then apply SVM v1 classifier to set 2, with the classification results verified in the same process as before. This time only 19 tracks were miss classified and needed individual annotations.

The process is repeated using truth labels from data sets 1 and 2 to train SVM v2, which is evaluated on data set 4 and showed only marginal performance improvement (row 3 in Table 1). SVM v2 is applied to data set 3 and verified in TAACKIT. Ten tracks were found to be misclassified and needed individual track annotation. Through this iterative process of ML training followed by annotation verification, we were able to experimentally determine that given the classification task, the data set and representation, and the algorithm choice, approximately 270 annotated tracks were sufficient to train the classifier.

TABLE 1. Performance Comparison in Model Iterations

| Model | Behavior | Metrics | | | |
|---|---|---|---|---|---|
| | | Accuracy | Precision | Recall | F1-Score |
| K-Means | Landing | 0.768 | 0.900 | 0.568 | 0.697 |
| | Touch&Go | | 0.295 | 0.474 | 0.364 |
| | Takeoff | | 0.907 | 0.986 | 0.944 |
| SVM v1 | Landing | 0.967 | 0.922 | 1.000 | 0.960 |
| | Touch&Go | | 0.789 | 0.882 | 0.882 |
| | Takeoff | | 0.993 | 0.993 | 0.993 |
| SVM v2 | Landing | 0.970 | 0.922 | 1.000 | 0.960 |
| | Touch&Go | | 1.000 | 0.816 | 0.899 |
| | Takeoff | | 1.000 | 0.993 | 0.996 |

*C. Annotation Effort*

To account for the effort required to label tracks with behavior classes using TAACKIT integrated with ML compared to a fully manual process, we propose the following accounting metric:

effort reduction = 1 – (number of single-track annotation + 6) / total number of tracks

The single-track annotations account for the misclassified tracks that need corrections. The 6 additional annotations account for the batch-annotation operations (2 runways * 3 behavior classes). The denominator is the equivalent fully manual annotation effort in number of tracks. Table 2 is an accounting of the annotation efforts in each iteration from our small-scale demonstration use case. The annotation effort reductions in our example use case range from 77% initially to 94% in the last iteration when compared with a fully manual annotation process that does not leverage ML.

TABLE 2. Annotation Effort Reduction

| Annot. cycle | Num of Tracks | Mis-classified | Annotation effort | Effort reduction |
|---|---|---|---|---|
| 1 | 271 | 57 | 63 | 77% |
| 2 | 273 | 19 | 25 | 91% |
| 3 | 271 | 10 | 16 | 94% |

We have demonstrated through this example a significant saving in the amount of effort required to generate annotated truth data in geospatial tracks to support air traffic behavior classifier development. Leveraging ML in the annotation process and TAACKIT, a ML-centric geospatial track annotation tool with built-in capability to ingest ML/analytics output, are crucial to the reduction of effort. While these numbers are measured from our specific use case, there is reason to believe that annotation effort reduction can be achieved in most cases by leveraging ML in the annotation process. The exact effort reduction will be dependent on specific application data and analytics goals.

## VI. SUMMARY

Geospatial track domain is a challenging area to apply machine learning techniques due to the lack of available and AI-ready datasets. We have introduced TAACKIT, a user interface and database tool, developed to enable annotation and validation of geospatial track ML development. We have demonstrated TAACKIT through a use case in an air traffic behavior analytics application to showcase the integration of annotation and analytics development to optimize ML processes. Leveraging iteratively and partially trained analytics models to automate annotation generation enabled us to transform annotation effort into verification tasks thus reducing overall annotation effort and speeding up overall development process. Integrating ML output in annotation process is possible in TAACKIT due to design considerations specifically targeted to support ML application development. We proposed a metric to quantify annotation effort and showed significant reduction in annotation effort of greater than 90%. This is a large reduction in annotation effort and can greatly improve the overall geospatial track analytics development speed. TAACKIT future development could include techniques such as active learning to prioritize annotation of "hard data", those data observations at the margin of misclassification or missed detections, that are most likely to improve analytics performance.

## ACKNOWLEDGMENT

The authors would like to thank MIT Lincoln Laboratory for sponsoring the development of TAACKIT. Additionally, we would like to thank Evan Maki for providing his expertise in interpreting the air traffic data in the use case demonstration.